\documentclass{ismrm}

\pdfoutput=1

\usepackage{makeidx}  
\usepackage{amsmath}
\usepackage{subfigure}
\usepackage{graphicx}
\usepackage{array}
\usepackage{bm}
\usepackage{amssymb}
\usepackage[%
  breaklinks=true,%
  colorlinks=true,%
  linkcolor=blue,anchorcolor=blue,%
  citecolor=blue,filecolor=blue,%
  menucolor=blue,pagecolor=blue,%
  urlcolor=blue]{hyperref}

\title{Statistical Noise Analysis in SENSE Parallel MRI}
\author{\large Santiago Aja-Fern\'{a}ndez\\
\normalsize LPI, ETSI Telecomunicaci\'on, Universidad de Valladolid, Spain\\
\normalsize Email: sanaja@tel.uva.es\\
\large Gonzalo Vegas-S\'anchez-Ferrero\\
\normalsize LPI, ETSI Telecomunicaci\'on, Universidad de Valladolid, Spain\\
\normalsize Email: gvegsan@lpi.tel.uva.es\\
\large Antonio Trist\'an-Vega\\
\normalsize LPI, ETSI Telecomunicaci\'on, Universidad de Valladolid, Spain\\
\normalsize Email: atriveg@lpi.tel.uva.es}

\runningtitle{Aja-Fern\'{a}ndez, Trist\'an-Vega, Vegas-S\'anchez-Ferrero}{LPI Tech Report: Statistical Noise Analysis in SENSE}
\wordcount{2794}

\begin{document}

\maketitle

\begin{abstract}
A complete first and second order statistical characterization of noise in SENSE reconstructed data is proposed. SENSE acquisitions have usually been modeled as Rician distributed, since the data reconstruction takes place into the spatial domain, where Gaussian noise is assumed. However, this model just holds for the first order statistics and obviates other effects induced by coils correlations and the reconstruction interpolation. Those effects are properly taken into account in this study, in order to fully justify a final SENSE noise model. As a result, some interesting features of the reconstructed image arise: (1) There is a strong correlation between adjacent lines. (2) The resulting distribution is non-stationary and therefore the variance of noise will vary from point to point across the image. Closed equations for the calculation of the variance of noise and the correlation coefficient between lines are proposed. The proposed model is totally compatible with $g$-factor formulations.
\end{abstract}


\section{Introduction}

An accurate statistical model of signal and noise is the keystone for many different applications in medical image processing and in the Magnetic Resonance (MR) field in particular. Traditionally, noise filtering techniques are based on a well-defined prior data statistical model. Many examples can be found in literature, such as the Conventional Approach~\cite{Mcgibney93}, ML~\cite{Sijbers98c} and LMMSE~\cite{AjaIP08,TristanMedia10,BrionMICCAI11} estimators or unbiased non-local mean filters~\cite{Wiest08,Manjon08,TristanNLM11}. A proper noise modeling may be useful not only for filtering purposes, but for many other processing techniques. Lately, for instance, Weighted Least Squares methods to estimate the Diffusion Tensor have proved to be nearly optimal when the data follows a Rician~\cite{Salvador05} or a non-central Chi (nc-$\chi$) distribution~\cite{Tristan09,TristanDWI11}. 

For practical purposes, the modeling is usually done assuming noise in MR data is a zero-mean spatially uncorrelated Gaussian process with equal variance in both the real and imaginary parts in each acquisition coil. As a result, in single coil systems magnitude data in the spatial domain are modeled using a stationary Rician distribution~\cite{Gudbjartsson95}. When multiple coils are considered and the {\bf k}-space is fully sampled, the natural extension of the Rician model yields to a stationary nc-$\chi$ distribution, whenever the different images are combined using sum of squares, the variance of noise is the same for all coils, and no correlations exist between them.
However, multiple coils systems are preferably set up to work with subsampled {\bf{k}}-space data, which have to be combined by some means to avoid the inherent aliasing artifact introduced by the under-completeness of the Fourier domain. The most popular algorithms to accomplish this task are GRAPPA~\cite{grappa} and SENSE~\cite{pruessmann99:sense}, the latter focusing the discussion in the present paper.

While the nc-$\chi$ has been used to describe noise in GRAPPA~\cite{Breuer09,AjaMRM11}, SENSE acquisitions have been usually modeled as Rician distributed, owing to the computation of the modulus of the complex signal linearly obtained from the array of coils~ \cite{Thunberg07}. However, this model stands exclusively for the first order statistics of noise, and obviates some other side effects induced by coils correlations and {\bf{k}}-space interpolation. For example, in GRAPPA reconstructions, both the initial inter-coil codependence and the {\bf{k}}-space interpolation introduce a strong correlation between the signals to be combined, and the nc-$\chi$ model is not strictly fulfilled. This inaccuracy can be worked around introducing an effective value of the power of noise and an effective number of receivers in the nc-$\chi$ distribution~\cite{AjaMRM11,AjaChi11}, both of them being spatially dependent.

For SENSE, the subsampling/interpolation effects have been previously described through the so-called $g$-factor, a global parameter that explicitly measures the SNR degradation in the acquisition process~\cite{Robson08}. 
In this work we aim at fully characterizing the first and second order statistics of noise in SENSE reconstructed images, including the effects of {\bf{k}}-space subsampling and inter-coil noise correlations. Since the reconstruction will take place into the spatial domain, and it can be seen as a weighted combination of the subsampled coils, the reconstructed image will be modeled as a complex Gaussian distribution. Its magnitude will be a non-stationary Rician distribution, with a spatial pattern that we can predict from certain imaging parameters such as the coils sensitivities and the speed-up factor (first order characterization). The study will show, in addition, another interesting feature of the reconstructed image: there exists a strong correlation between adjacent lines in the reconstructed volume, with an extent directly dependent on the acceleration factor (second order characterization).

\section{Theory}

\subsection{Statistical Model of MR signals}

The {\bf{k}}-space data at each coil of the MR scanner can be accurately described by a noise-free signal plus an Additive White Gaussian Noise (AWGN) process, with zero mean and variance $\sigma_{K_l}^2$:
\begin{equation}
s_l({\bf k})={a_l}({\bf k})+n_l({\bf k};0,\sigma_{K_l}^2),\ \ \ l=1,\cdots,L
\end{equation}
with $a_l({\bf k})$ the noise-free signal and $n_l({\bf k};0,\sigma_{K_l}^2)=n_{l_r}({\bf k};0,\sigma_{K_l}^2)+jn_{l_i}({\bf k};0,\sigma_{K_l}^2)$ the AWGN process, which is initially assumed stationary so that $\sigma_{K_l}^2$ does not depend on {\bf k}.
The complex {\bf{x}}-space is obtained as the inverse Discrete Fourier Transform (iDFT) of $s_l(\mathbf{k})$ for each slice or volume, so the noise in the complex {\bf{x}}-space is still assumed to be Gaussian:
$$
S_l({\bf x})=A_l({\bf x})+N_l({\bf x};0,\sigma_l^2), \ \ \ l=1,\cdots,L
$$
where $N_l({\bf x};0,\sigma_l^2)=N_{l_r}({\bf x};0,\sigma_l^2)+jN_{l_i}({\bf x};0,\sigma_l^2)$ is also a complex AWGN process (assuming that there are not any spatial correlations) with zero mean and covariance matrix ${\bf \Sigma}$: 
$$
{\bf \Sigma} =\left(\begin{array}{cccc} 
\sigma_1^2& \sigma_{12}^2 &\cdots& \sigma_{1L}^2\\
\sigma_{21}^2& \sigma_2^2& \cdots& \sigma_{2L}^2\\
\vdots& \vdots &\ddots &\vdots\\
\sigma_{L1}^2& \sigma_{L2}^2& \cdots &\sigma_L^2
\end{array}\right),
$$
The variance of noise for each coil in {\bf k}- and {\bf x}-spaces are related through the number of points in the image:
\begin{equation}
\sigma_l^2=\frac{1}{|\Omega|}\sigma_{K_l}^2
\label{eq:xkrelation}
\end{equation}
with $|\Omega|$ the size of the image in each coil, i.e. the number of points used in the 2D iDFT.
If the {\bf k}-space is fully sampled, the Composite Magnitude Signal (CMS) can be directly obtained using SoS~\cite{Constantinides97,Roemer90}:
\begin{equation}
M_L(\mathbf x)=\sqrt{\sum_{l=1}^L|S_{l}(\mathbf x)|^2 }.
\label{eq:parallelmodel}
\end{equation}
For a single--coil acquisition, the CMS, $M({\bf x})$, is the Rician distributed envelope of the complex signal~\cite{MRM:Gudbjartsson:95}. In the image background, where the signal-to-noise ratio is zero due to the lack of water-proton density in the air, the Rician simplifies to a Rayleigh distribution. For multiple coils, if the variance of noise is the same for all coils, no correlation exists between them, and the signals are combined using SoS, the CMS may be modeled as a nc-$\chi$ distribution~\cite{Constantinides97,Dietrich08,AjaMRI09,AjaMRM11,Koay06}. In a more general case where correlations are taken into account, the nc-$\chi$ is only an approximation of the real distribution. It can be accurately approximated with this model if effective parameters (reduced number of coils and and increased variance of noise) are used~\cite{AjaChi11}.


\subsection{Statistical model in SENSE reconstructed images}
\label{sec:paral}

\begin{figure}[tb]
\centering
\includegraphics{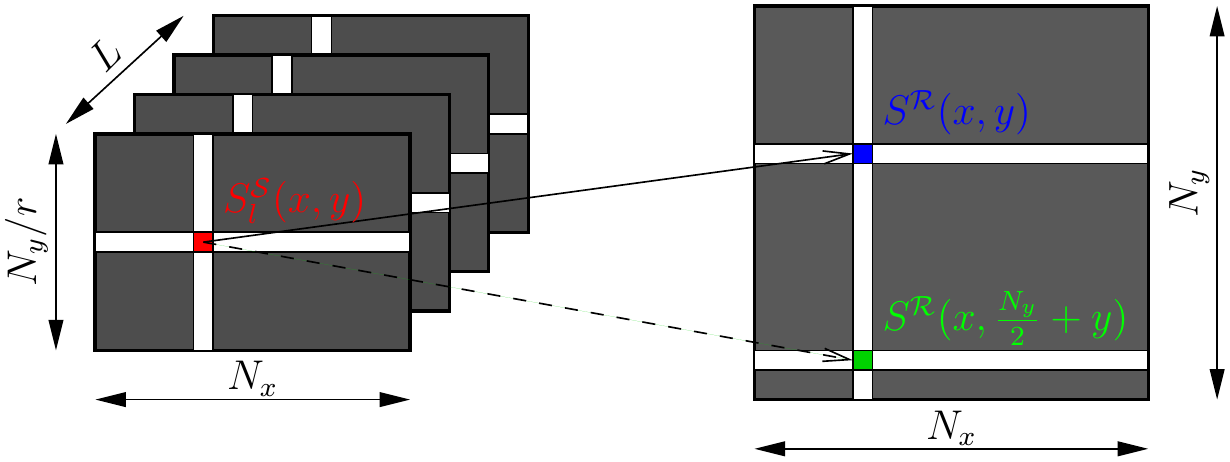}
\caption{Example of the SENSE interpolation for 4 coils and an acceleration factor $r=2$. }
\label{imag:ejemplo}
\end{figure}

For the sake of simplicy, let us assume that SENSE~\cite{pruessmann99:sense} is only be applied to MRI data regularly subsampled by a factor $r$. The reconstruction takes place in the image domain. Assuming an original size $|\Omega|=M_x \times M_y$, the subsampled signal in the {\bf x}-space ${S}^{\mathcal S}_l({\bf x})={S}^{\mathcal S}_l(x,y)$ is the (complex) Fourier inverse transform of ${s}^{\mathcal S}_l({\bf k})$, of size $M_x \times (M_y/r)$ . Note than since the size of subsampled data in each coil in the {\bf x}-space is reduced by a factor $r$, the variance of noise will be umplified by that same factor:
$$
\sigma_l^2=\frac{r}{|\Omega|}\sigma_{K_l}^2
$$

\begin{figure}[tb]
\centering
\includegraphics{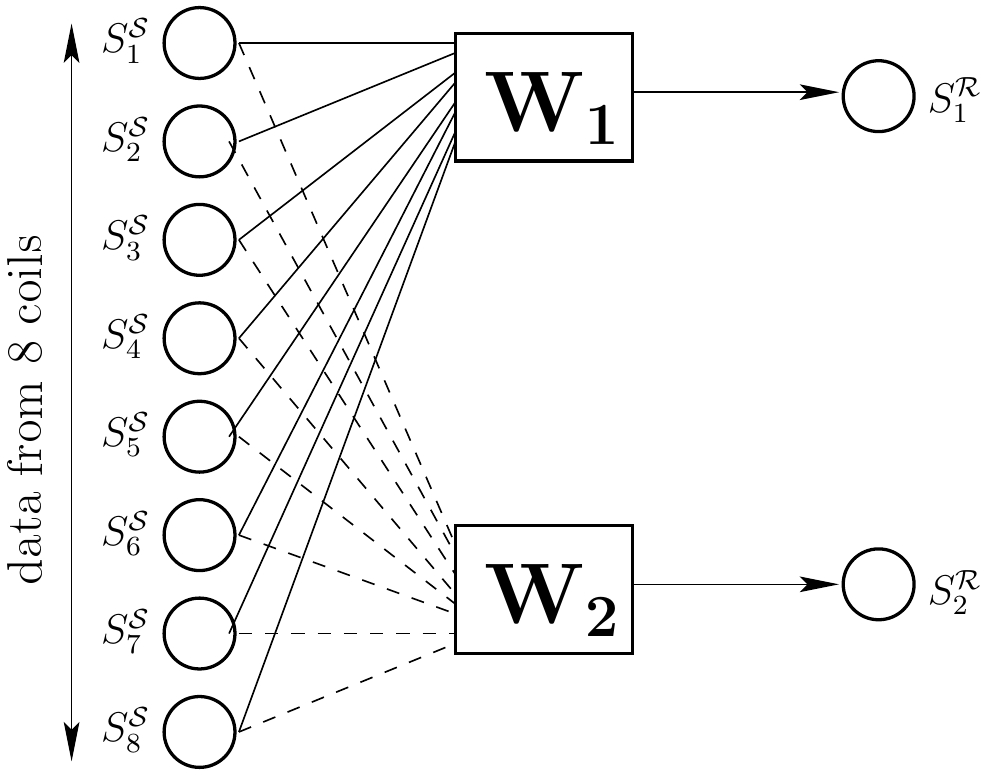}
\caption{Example of the SENSE interpolation for 8 coils and an acceleration factor $r=2$. }
\label{imag:scheme}
\end{figure}

In multiple coil scanners, the image received in coil $l$-th, $S_l(x,y)$, can be seen as an {\em original image} $S_0(x,y)$ weighted by the sensitivity of that specific coil:
\begin{equation}
S_l(x,y)=C_l(x,y)S_0(x,y), \ \ \ l=1,\cdots,L
\end{equation}
An accelerated pMRI acquisition with a factor $r$ will reduce the matrix size of the image at every coil. The signal in one pixel at location $(x,y)$ of $l$-th coil can be now written as~\cite{Blaimer04}:
\begin{equation}
S_l(x,y)=C_l(x,y_1)S_0(x,y_1)+\cdots+C_l(x,y_r)S_0(x,y_r)
\label{eq:rec1}
\end{equation}
In SENSE, the reconstructed image ${S}^{\mathcal R}(x,y)$ can be seen as an estimator of the original image ${S}^{\mathcal R}(x,y)=\widehat{S_0}(x,y)$ that can be obtained from eq.~(\ref{eq:rec1}). For instance, for $r=2$ for pixel $(x,y)$, ${S}^{\mathcal R}(x,y)$ is obtained as
\begin{equation}
\left[\begin{array}{c}
S^{\mathcal R}_1\\
S^{\mathcal R}_2
\end{array}\right] =\left[\begin{array}{cc}
{\bf W}_1&{\bf W}_2
\end{array}\right]\times\left[\begin{array}{ccc}
S^{\mathcal S}_1&\cdots &S^{\mathcal S}_L
\end{array}\right].
\end{equation}
In matrix form:
\begin{equation}
{\bf S}^{\mathcal R}={\bf W}\times{\bf S}^{\mathcal S}
\end{equation}
with ${\bf W}(x,y)=\left[{\bf W}_1,  \cdots,  {\bf W}_r \right]$ a reconstruction matrix created from the sensitivity map of each coil, ${\bf C}(x,y)=[{\bf C}_1, \cdots, {\bf C}_l]$:
$$
{\bf W}(x,y)=({\bf C}^*(x,y){\bf C}(x,y))^{-1}{\bf C}^*(x,y)
$$
If the correlation between coils is taken into account, the reconstruction matrix must incorporate the covariance matrix: 
$$
{\bf W}(x,y)=({\bf C}^*(x,y){\bf \Sigma}^{-1}{\bf C}(x,y))^{-1}{\bf C}^*(x,y){\bf \Sigma}^{-1}
$$
For the sake of simplicity, we will remove any pixel dependency, so that we can write for each output pixel:
\begin{equation}
S^{\mathcal R}_i={\bf W}_{i} \times{\bf S}^{\mathcal S}\ \ \ i=1,\cdots ,r
\label{eq:sense:simplified}
\end{equation}
Two examples can be found on Fig.~\ref{imag:ejemplo} and on Fig.~\ref{imag:scheme}.

The SNR of the fully sampled image and the image reconstructed with SENSE are related by the so-called g-factor, $g$~\cite{Robson08,Blaimer04}:
\begin{equation}
\mbox{SNR}_{\mbox{\tiny SENSE}}=\frac{\mbox{SNR}_{\mbox{\tiny full}}}{\sqrt{r}\cdot g}
\end{equation}

However, we will focus on the actual noise model underlying the SENSE reconstruction and on the final variance of noise. The final signal $S^{\mathcal R}_i$ is obtained as a linear combination of $S^{\mathcal S}_l$, where the noise is Gaussian distributed. Thus, the resulting signal is also Gaussian, with variance:
\begin{equation}
\sigma_i^2={\bf W}_{i}^* {\bf \Sigma} {\bf W}_{i}
\label{eq:var:noise}
\end{equation}
Since ${\bf W}_{i}$ is position dependent, i.e.  ${\bf W}_{i}={\bf W}_{i}(x,y)$, so will be the variance of noise, $\sigma_i^2(x,y)$. For further reference, when the whole image is taken into account, let us denote the variance of noise for each pixel in the reconstructed data by  $\sigma_{\mathcal R}^2({\bf x})$.

Note now that all the lines $S^{\mathcal R}_i$ reconstructed from the same data ${S}_l^{\mathcal S}$ will be strongly correlated, since they are basically different linear combinations of the same Gaussian variables. In that case, the covariance between $S^{\mathcal R}_i$ and $S^{\mathcal R}_j$, $i\neq j$ can be calculated as
\begin{equation}
\sigma_{i,j}^2={\bf W}_{i}^* {\bf \Sigma} {\bf W}_{j}
\end{equation}
and the correlation coefficient is derived straight forward:
\begin{equation}
\rho_{i,j}^2=\frac{\sigma_{i,j}^2}{\sigma_{i}\sigma_{j}}=\frac{{\bf W}_{i}^* {\bf \Sigma} {\bf W}_{j}}{\sqrt{\left({\bf W}_{i}^* {\bf \Sigma} {\bf W}_{i}\right)\left({\bf W}_{j}^* {\bf \Sigma} {\bf W}_{j}\right)}},
\label{eq:corr:coef}
\end{equation}
However, these correlations are not strongly affecting the data, since the correlated pixels are separated by $N_y/r$ lines

All in all, noise in the final reconstructed signal $S^{\mathcal R}(x,y)$ will follow a complex Gaussian distribution. If the magnitude is considered, i.e. $M(x,y)=|S^{\mathcal R}(x,y)|$, the final CMS will follow a Rician distribution, just like single-coil systems.

We can summarize our developments as follows:
\begin{enumerate}
\item Subsampled multi coil MR data reconstruted with cartesian SENSE follows a Rician distribution in each point of the image.
\item The resulting distribution is non-stationary. This means that the variance of noise will vary from point to point across the image.
\item The variance of noise final value in each point will only depend on the covariance matrix of the original data and on the sensitivity map. 
\item Each pixel in the final image will be strongly correlated with all those pixels reconstructed from the same original data. Each pixel is correlated with $r-1$ other pixels. These correlated pixels are far enough and they can be neglected. 
\end{enumerate}

For the particular case in which there is no correlation between coils and all the coils has the same noise variance $\sigma_n^2$, we can write eq.~(\ref{eq:var:noise}) as:
\begin{equation}
\sigma_i^2=\sigma_n^2 \times |{\bf W}_{i}|^2
\end{equation}
Since $\sigma_n^2$ is the noise variance for the subsampled data in the {\bf-x}-space, according to eq.~(\ref{eq:xkrelation}), it is related to the original noise level without subsampling, say $\sigma_0^2$, by the subsampling rate:
$$
\sigma_n^2=r\cdot \sigma_0^2
$$
and therefore
\begin{equation}
\sigma_i=\sqrt{r}\cdot \sigma_0 \times |{\bf W}_{i}|
\end{equation}
which is totally equivalent to the g-factor formulations for SNR reduction in literature \cite{pruessmann99:sense,Thunberg07}.

\section{Materials and Methods}

For the sake of validation and illustration of the results in the previous section, the following experiments are considered:

{\bf First}, we will study the statistical behavior of Gaussian data when a combination like the one in SENSE is done. To that end, we consider $10^5$ samples of 8 correlated complex Gaussian RVs with zero mean and unitary variance, $N(0,1)$ and two different correlation coefficients, $\rho^2=0$ and $\rho^2=0.2$. (Note the correlation is between variables, not between samples of the same variable). The RVs are combined using real random weights, ${\bf W}_1$ and ${\bf W}_2$, both following a uniform distribution in $[0,1]$ and normalized so that
$$
|{\bf W}_i|^2=1,\ \ \ i=1,2
$$
Two new variables are created by using a combination like the one in eq.~(\ref{eq:sense:simplified}), obtaining two new RVs. The sample variance and correlation coefficient are estimated from the data and then compared to those calculated from eq.~(\ref{eq:var:noise}) and eq.~(\ref{eq:corr:coef}).

\begin{figure}[tb]
\centering
\subfigure{\includegraphics[width=0.9\textwidth]{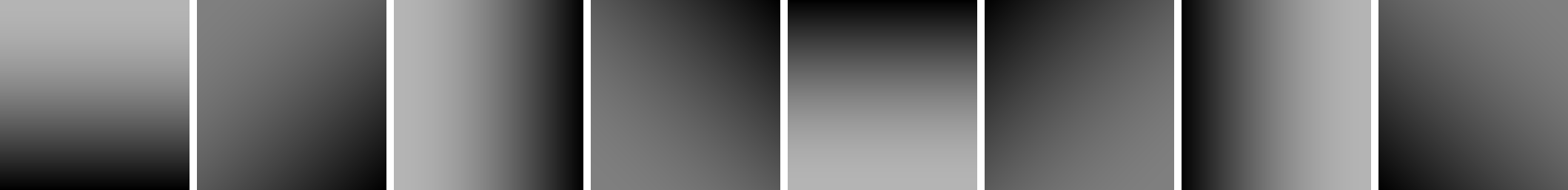}}\hfil
\subfigure{\includegraphics[width=0.9\textwidth]{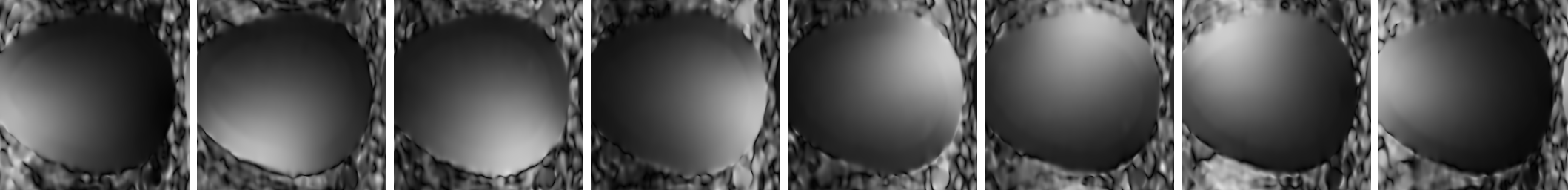}}
\caption{Sensitivity Maps used for the experiment. Top: Synthetic sensitivity map created so that the SoS of the maps gives a constant image. Bottom: Sensitivity map estimated from an actual brain imaging acquisition in a SENSE Signa 1.5T scanner with 8 coils.}
\label{imag:sensitivity}
\end{figure}

{\bf Secondly}, we will test how the values of $\sigma_{\mathcal R}^2({\bf x})$ varies across the image. To that end, we will work with one sensitivity map synthetically generated, as shown in Fig.~\ref{imag:sensitivity} (top). This map simulates an 8-coil system using an artificial sensitivity map coded for each coil so that $\sum_{l} |C_l({\bf x})|^2=1$, with $l=1\cdots, 8$, and $C_l({\bf x})$ the sensitivity map of coil $l$-th. For the experiment:
\begin{itemize}
\item We assume that each coil has an original variance of noise $\sigma^2_l=100$. We will simulate two different configurations, first, assuming that there is no initial correlation between coils, and second, assuming a correlation coefficient of $\rho^2=0.1$ between all coils, so that
$$
{\bf \Sigma} =100\times\left(\begin{array}{cccc} 
1& 0.1 &\cdots& 0.1\\
0.1& 1& \cdots& 0.1\\
\vdots& \vdots &\ddots &\vdots\\
0.1& 0.1& \cdots &1
\end{array}\right).
$$
\item From the data, and using the theoretical expressions in eq.~(\ref{eq:var:noise}) and eq.~(\ref{eq:corr:coef}) we calculate the variance of noise for each pixel in the final image.
\item In order to test the theoretical distributions, 5000 samples of 8 complex $256\times 256$ Gaussian images with zero mean and covariance matrix ${\bf \Sigma}$ are generated. The {\bf k}-space of the data is subsampled by a 2x factor and reconstructed using SENSE and the synthetic sensitivity field. We estimate the variance of noise in each point using the second order moment of the Rayleigh distribution \cite{AjaMRI09}:
$$
\sigma^2_{\mathcal R}({\bf x})=\frac{1}{2}E\{M^2({\bf x})\}
$$
\end{itemize}
We estimate the $E\{M^2({\bf x})\}$ along the 5000 samples. 

{\bf Last}, the previous experiment is repeated for the correlated case, now using real sensitivity maps as shown in Fig.~\ref{imag:sensitivity} (bottom). These maps are estimated from a real brain T1 acquisition  done in a GE Signa 1.5T EXCITE, FSE pulse sequence, 8 coils, TR=500msec, TE=13.8msec, image size $256\times 256$ and FOV: 20cm$\times$20cm.

\section{Results}

\begin{table}
\begin{center}
\begin{tabular}{|l|l|c|c|}
\hline
Experiment&Parameter & Sample Value & Theoretical\\                 
\hline
$\rho^2=0 $  & $\sigma_1$  &	1.0072  & 1.0000  \\
        & $\sigma_2$   &  1.0023  & 1.0000 \\
        &$\rho_{1,2}^2$&	0.9262 & 0.9247\\
\hline
$\rho^2=0.2$ & $\sigma_1$	& 1.4198 &1.4121 \\
           &    $\sigma_2$  &  1.4536  & 1.4412 \\
            &  $\rho_{1,2}^2$	& 0.9706 &0.9703\\
\hline
\end{tabular}
\end{center}
\caption{Results from the first experiment. Standard deviation and correlation coefficient of the SENSE-like combination of synthetic Gaussian data. Theoretical and sample values.}
\label{table:exp1}
\end{table}

The {\bf first} experiment studies the behavior of a SENSE-like combination of Gaussian data. Results of the standard deviation and correlation for the two resulting variables are collected in Table~\ref{table:exp1}. As expected, the theoretical values match the estimation through samples. Note that the variances in the final RVs in the correlated case are higher than the ones in the case without correlation. This effect can be found in real data, where correlations exist and must be taken into account.

\begin{figure}[tb]
\centering
\subfigure{\includegraphics[width=0.33\textwidth]{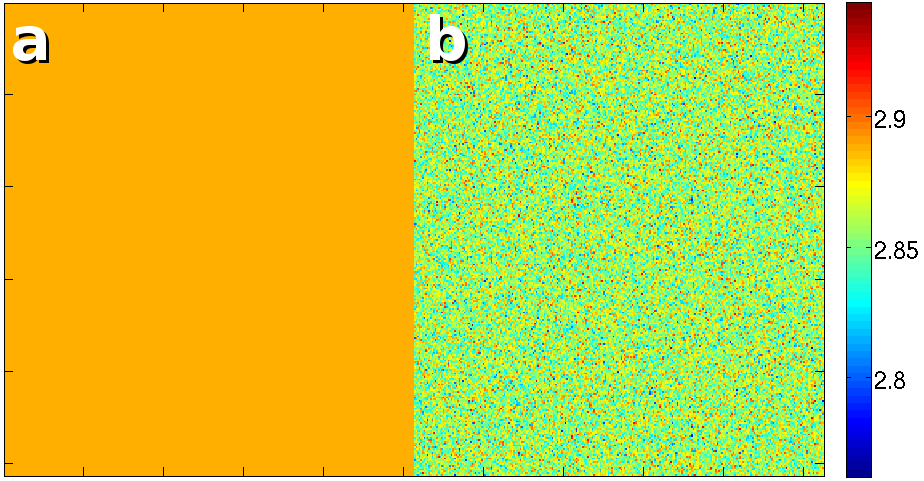}}\hfil
\subfigure{\includegraphics[width=0.33\textwidth]{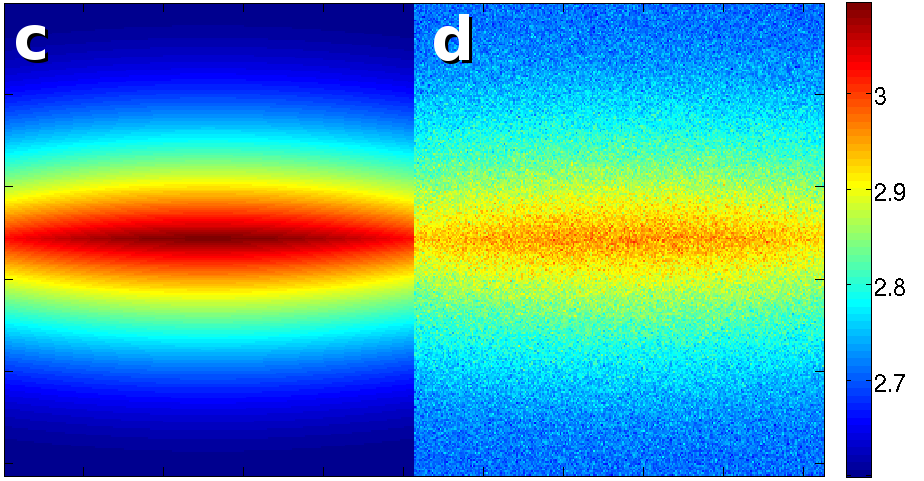}}\hfil
\subfigure{\includegraphics[width=0.31\textwidth]{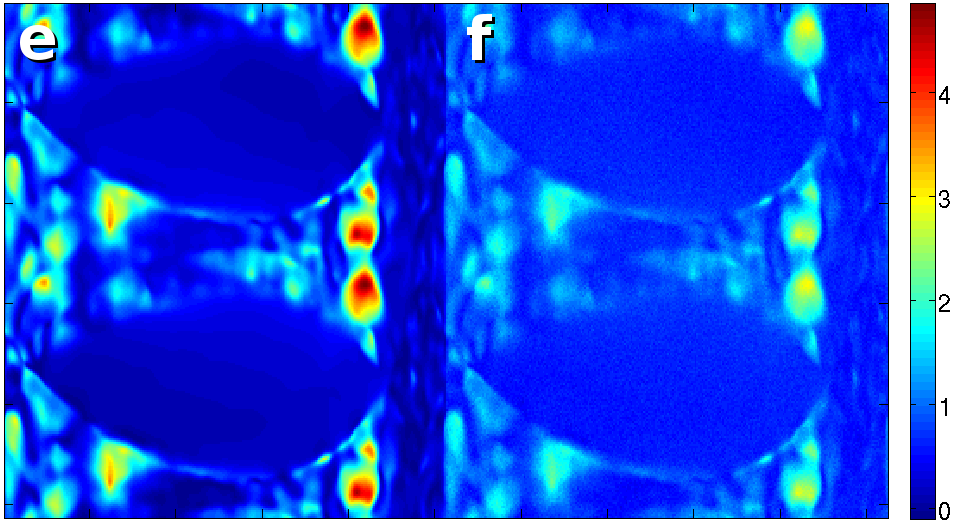}}\
\caption{Maps of standard deviation of noise $\sigma_{\mathcal R}({\bf x})$ in the final image: (a-c-e): Theoretical values. (b-d-f): Estimated from samples. (a-b) Synthetic Sensitivity Map with no correlation. (c-d) Synthetic Sensitivity Map with correlation between coils. (e-f) Real sensitivity map with correlation between coils.}
\label{imag:sensitivity2}
\end{figure}

The {\bf second} and {\bf third} experiments deals with the non homogeneous spatial layout of the noise and the influence of the sensitivity map over the final noise distribution. Visual results are depicted in Fig~\ref{imag:sensitivity2}. For the synthetic maps, when no correlations are considered, since $\sum_{l} |C_l({\bf x})|^2=1$ for all pixels, the final variance of noise will not depend on the position ${\bf x}$. Therefore, in this particular case $\sigma^2_{\mathcal R}({\bf x})=\sigma^2_{\mathcal R}$. The estimated values in Fig~\ref{imag:sensitivity2}-(b) show a noise pattern that slightly varies around the real value (note the small range of variation). In this very particular case, the noise can be considered to be spatially stationary, and the final image (leaving the correlation between pixels aside) is equivalent to one obtained from a single-coil scanner.

When correlations are taken into account, even using the same synthetic sensitivity map, results differ. In Fig.~\ref{imag:sensitivity2}-(c), the theoretical value shows that the standard deviation of noise of the reconstructed data is not the same for every pixel, i.e., the noise is no longer spatial-stationary. The center of the image shows a larger value that decreases going north and south. So, in this more realistic case, the $\sigma^2_{\mathcal R}({\bf x})$ will depend on ${\bf x}$, which can have serious implications for future processing, such as model based filtering techniques. The estimated value in Fig.~\ref{imag:sensitivity2}-(d) shows exactly the same non-homogeneous pattern across the image. 

In the last experiment, Fig.~\ref{imag:sensitivity2}-(e) and Fig.~\ref{imag:sensitivity2}-(f), a real sensitivity map is used, and correlation between coils is also assumed. Again, the noise is non-stationary. To increase the dynamic range of the images, the logarithm has been used to show the data.

\section{Discussion and conclusions}

Noise analysis in SENSE in literature is usually focused on the study of the SNR loss due to the acceleration process, being the so-called g-factor the most common measure. However, despite its proved utility, the g-factor is insufficient when trying to design certain MR applications. In those cases, the need of a complete statistical modeling arises naturally. In SENSE, the Rician model has been widely assumed by the MR community, as seen in recent literature~\cite{Dietrich08,Thunberg07}. This assumption has traditionally been taken as a guarantee to use the methods designed for single coil MR data over multiple-coil SENSE reconstructed data.

The study of the noise distribution in SENSE carried out in this work has brought to light some serious implications that must have been taken into account when working with SENSE data. Even when the final distribution is always Rician, depending on the sensitivity maps and on the coils covariance matrix, this distribution is likely to be non-stationary. In the examples proposed, even in the optimal synthetic case, when correlations between coils were present, the non-stationarity arises. As a consequence, the variance of noise will differ from point to point across the image. The first implication of this feature is related with noise estimation. Since the variance of noise depends on the position, most noise estimation techniques, as the ones proposed in \cite{AjaMRI09}, have no longer sense. There is no longer a single value to estimate for the whole image, but one for each point. Thus, those algorithms that need an estimation of the level of noise, cannot be used either in its original shape. See, for instance, those filtering techniques reviewed in the Introduction.

%
As an example, let us consider a very simple noise reduction technique, the Conventional Approach (CA) for Rician data~\cite{Mcgibney93}:
$$
\widehat{A}({\bf x})=\sqrt{E\{M^2_L({\bf x})\}-2\sigma_n^2},
$$
whose same philosophy is shared by several other methods mainly based on NLM denoising~\cite{Wiest08,Manjon08,TristanNLM11}. When dealing with SENSE reconstructions, the variance of noise $\sigma_n^2$ will no longer be unique for the whole image, i.e. $\sigma_n^2({\bf x})$. The noise estimation is no longer a simple task. Some prior data {\em regulation} like the one done in \cite{AjaISBI11} could be necessary.
%
Note that, even an approach as simple as the CA cannot be directly applied from the single coil Rician formulation over SENSE data.

To sum up, SENSE MR data is known to follow a Rician distribution, but non-homogeneity of the variance of noise arises due to the reconstruction process. A prior knowledge of the sensitivity maps and the coils covariance matrix will help to properly design applications  to deal with this kind of data.

\section*{Acknowledgments}
The authors acknowledge Ministerio de Ciencia e Innovaci\'on for grant TEC2010-17982. The sensitivity maps used were kindly provided by Doctor W. Scott Hoge from the LMI, Brigham and Women’s Hospital, Boston.


\end{document}